\documentclass[a4paper]{article}
\usepackage{arxiv}
\usepackage{times}
\usepackage{hyperref}
\usepackage{multirow}
\usepackage{graphicx}
\usepackage{amsmath,amssymb,amsbsy,amsfonts,amsthm}
\makeatletter
\renewcommand*{\eqref}[1]{%
  \hyperref[{#1}]{\textup{\tagform@{\ref*{#1}}}}%
}
\makeatother
\usepackage[utf8]{inputenc}
\usepackage{biblatex}
\AtEveryBibitem{
	\clearlist{language}
	\clearfield{doi}
	\clearfield{issn}
	\clearfield{isbn}
	\clearfield{urlyear}
	\ifentrytype{misc}{}{\clearfield{url}}
}
\usepackage{subcaption}
\usepackage{tikz}
\usetikzlibrary{arrows,arrows.meta,calc,patterns,decorations.pathmorphing,backgrounds,positioning,fit}
\usepackage{array}

\newcommand{\tuple}[1]{\langle{#1}\rangle}
\tikzset{
n/.style={draw,rectangle,anchor=west},
t/.style={inner sep=0,anchor=west},
defbox/.style={rounded corners=2mm,draw=black!30,line width=1mm},
cartouche/.style={fill=black!40,line width=1mm,anchor=south},
past/.style={thick,blue},
future/.style={thick,black!40!green},
nproc/.style={thick,dashed},
tproc/.style={thick,dotted}
}
\title{Structured Time Series Prediction\\without Structural Prior}
\renewcommand{\shorttitle}{Structured Time Series Prediction without Structural Prior}
\renewcommand{\headeright}{}
\renewcommand{\undertitle}{}
\author{Darko Drakulic, Jean-Marc Andreoli\\NAVER LABS Europe, Grenoble, France\\
{\small\tt http://www.europe.naverlabs.com}}
\date{September 2021}
\begin{document}

\maketitle

\begin{abstract}
Time series prediction is a widespread and well studied problem with applications in many domains (medical, geoscience, network analysis, finance, econometry etc.). In the case of multivariate time series, the key to good performances is to properly capture the dependencies between the variates. Often, these variates are structured, i.e. they are localised in an abstract space, usually representing an aspect of the physical world, and prediction amounts to a form of diffusion of the information across that space over time. Several neural network models of diffusion have been proposed in the literature.
However, most of the existing proposals rely on some a priori knowledge on the structure of the space, usually in the form of a graph weighing the pairwise diffusion capacity of its points. We argue that this piece of information can often be dispensed with, since data already contains the diffusion capacity information, and in a more reliable form than that obtained from the usually largely hand-crafted graphs. We propose instead a fully data-driven model which does not rely on such a graph, nor any other prior structural information. We conduct a first set of experiments to measure the impact on performance of a structural prior, as used in baseline models, and show that, except at very low data levels, it remains negligible, and beyond a threshold, it may even become detrimental. We then investigate, through a second set of experiments, the capacity of our model in two respects: treatment of missing data and domain adaptation.
\end{abstract}

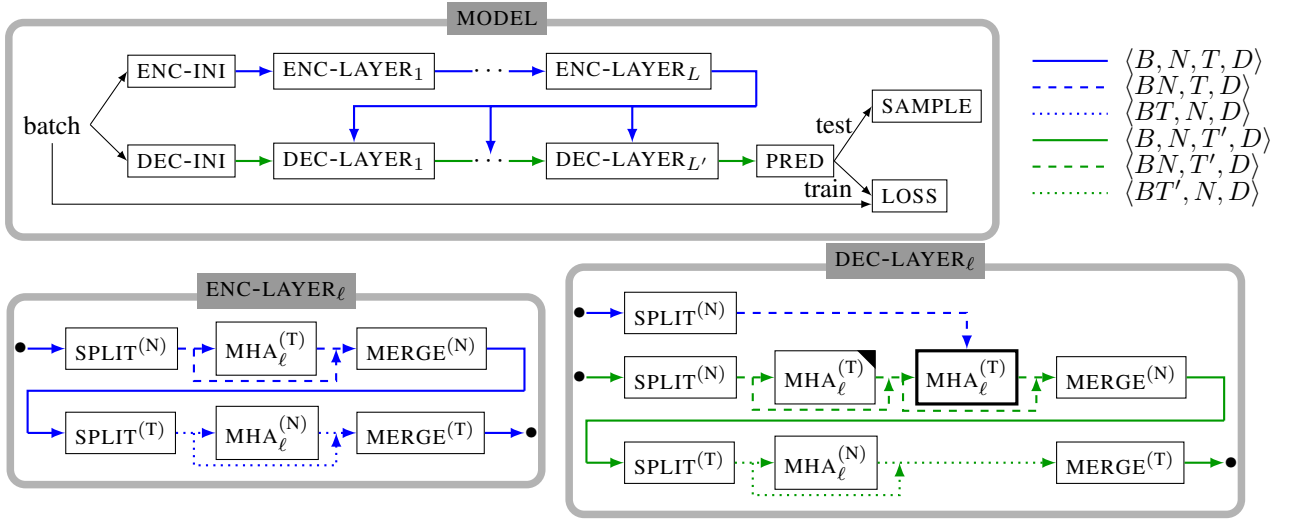
\begin{figure*}
\begin{center}
\begin{subfigure}{1.1\textwidth}
\begin{tikzpicture}[scale=.7]
\begin{scope}[local bounding box=bbox]
\matrix[column sep=5mm,row sep=0mm,anchor=north west,ampersand replacement=\&] {
\& 
\node[n] (enc0) {\sc enc-ini}; \&
\node[n] (enc1) {\sc enc-layer$_1$}; \&
\node[t] (encD) {$\cdots$}; \&
\node[n] (enc2) {\sc enc-layer$_L$};
\\
\node (batch) {batch}; \& \& \& \& \& \coordinate (enc-dec); \& \node[n] (sample) {\sc sample};
\\
\& 
\node[n] (dec0) {\sc dec-ini}; \&
\node[n] (dec1) {\sc dec-layer$_1$}; \&
\node[t] (decD) {$\cdots$}; \&
\node[n] (dec2) {\sc dec-layer$_{L'}$}; \&
\node[n] (pred) {\sc pred}; \&
\\
\& \& \& \& \& \& \node[n] (loss) {\sc loss};
\\
};
\draw[-latex] (batch.east) -- (enc0.west);\draw[-latex] (batch.east) -- (dec0.west);
\draw[-latex,past] (enc0) -- (enc1);\draw[past] (enc1) -- (encD);\draw[-latex,past] (encD) -- (enc2);
\draw[past] (enc2) -| (enc-dec);\draw[-latex,past] (enc-dec) -| (dec1);\draw[-latex,past] (enc-dec) -| (decD);\draw[-latex,past] (enc-dec) -| (dec2);
\draw[-latex,future] (dec0) -- (dec1);\draw[future] (dec1) -- (decD);\draw[-latex,future] (decD) -- (dec2);\draw[-latex,future] (dec2) -- (pred);
\draw[-latex] (pred.east) -- node [midway,above left,inner sep=0] {test} (sample.west);\draw[-latex] (pred.east)  -- node [midway,below left,inner sep=0] {train} (loss.west);
\draw[-latex] (batch) |- ($(loss.west)!.5!(loss.south west)$);
\end{scope}
\draw[defbox] ($(bbox.north west)+(.1,.3)$) rectangle ($(bbox.south east)+(.1,-.3)$);
\node[cartouche] at (bbox.north) {\sc model};
\begin{scope}[shift={($(bbox.north east)+(.6,0)$)}]
\matrix[column sep=2mm,row sep=0mm,anchor=north west,ampersand replacement=\&] {
\draw[past] ++(-.5,0) -- (.5,0); \& \node[t] {$\tuple{B,N,T,D}$};\\
\draw[past,nproc] ++(-.5,0) -- (.5,0); \& \node[t] {$\tuple{BN,T,D}$};\\
\draw[past,tproc] ++(-.5,0) -- (.5,0); \& \node[t] {$\tuple{BT,N,D}$};\\
\draw[future] ++(-.5,0) -- (.5,0); \& \node[t] {$\tuple{B,N,T',D}$};\\
\draw[future,nproc] ++(-.5,0) -- (.5,0); \& \node[t] {$\tuple{BN,T',D}$};\\
\draw[future,tproc] ++(-.5,0) -- (.5,0); \& \node[t] {$\tuple{BT',N,D}$};\\
};
\end{scope}

\end{tikzpicture}
\end{subfigure}
\\
\begin{subfigure}{0.35\textwidth}
\begin{tikzpicture}[scale=.7]
\begin{scope}[local bounding box=bbox]
\matrix[column sep=5mm,row sep=2mm,anchor=north west,ampersand replacement=\&] {
\node[t,anchor=east] (in) {$\bullet$}; \&
\node[n] (split-N) {{\sc split}$^{(\textrm{N})}$}; \&
\node[n] (mha-N) {{\sc mha}$_\ell^{(\textrm{T})}$}; \&
\node[n] (merge-N) {{\sc merge}$^{(\textrm{N})}$};
\\
\& \& \& \& \coordinate (N-T);
\\
\coordinate (N-T-);\&
\node[n] (split-T) {{\sc split}$^{(\textrm{T})}$}; \&
\node[n] (mha-T) {{\sc mha}$_\ell^{(\textrm{N})}$}; \&
\node[n] (merge-T) {{\sc merge}$^{(\textrm{T})}$}; \&
\node[t] (out) {$\bullet$};
\\
};
\draw[-latex,past] (in) -- (split-N);\draw[-latex,past,nproc] (split-N) -- (mha-N);\draw[-latex,past,nproc] (mha-N) -- (merge-N);\draw[past] (merge-N) -| (N-T);\draw[past] (N-T) -| (N-T-);
\draw[-latex,past,nproc] ($(split-N)!.5!(mha-N)$) |- ($(mha-N.south)+(0,-.1)$) -| ($(mha-N)!.45!(merge-N)$);
\draw[-latex,past] (N-T-) -- (split-T);\draw[-latex,past,tproc] (split-T) -- (mha-T);\draw[-latex,past,tproc] (mha-T) -- (merge-T);\draw[-latex,past] (merge-T) -- (out);
\draw[-latex,past,tproc] ($(split-T)!.5!(mha-T)$) |- ($(mha-T.south)+(0,-.1)$) -| ($(mha-T)!.45!(merge-T)$);
\end{scope}
\draw[defbox] ($(bbox.north west)+(.1,.3)$) rectangle ($(bbox.south east)+(-.1,-.3)$);
\node[cartouche] at (bbox.north) {\sc enc-layer$_\ell$};

\end{tikzpicture}
\end{subfigure}
\hfill
\begin{subfigure}{0.55\textwidth}
\begin{tikzpicture}[scale=.7]
\begin{scope}[local bounding box=bbox]
\matrix[column sep=5mm,row sep=2mm,anchor=north west,ampersand replacement=\&] {
\node[t,anchor=east] (in1) {$\bullet$}; \&
\node[n] (split1-N) {{\sc split}$^{(\textrm{N})}$}; \&
\\
\node[t,anchor=east] (in) {$\bullet$}; \&
\node[n] (split-N) {{\sc split}$^{(\textrm{N})}$}; \&
\node[n] (mha-N-self) {{\sc mha}$_\ell^{(\textrm{T})}$}; \fill[black] (mha-N-self.north east) -- ++(-.25,0) -- ++(.25,-.25) -- ++(0,.25) -- cycle;\&
\node[n,very thick] (mha-N-cross) {{\sc mha}$_\ell^{(\textrm{T})}$}; \&
\node[n] (merge-N) {{\sc merge}$^{(\textrm{N})}$};
\\
\& \& \& \& \& \coordinate (N-T);
\\
\coordinate (N-T-);\&
\node[n] (split-T) {{\sc split}$^{(\textrm{T})}$}; \&
\node[n] (mha-T) {{\sc mha}$_\ell^{(\textrm{N})}$}; \&
\&
\node[n] (merge-T) {{\sc merge}$^{(\textrm{T})}$}; \&
\node[t] (out) {$\bullet$};
\\
};
\draw[-latex,past] (in1) -- (split1-N); \draw[-latex,past,nproc] (split1-N) -| (mha-N-cross);
\draw[-latex,future] (in) -- (split-N);\draw[-latex,future,nproc] (split-N) -- (mha-N-self);\draw[-latex,future,nproc] (mha-N-self) -- (mha-N-cross);\draw[-latex,future,nproc] (mha-N-cross) -- (merge-N);\draw[future] (merge-N) -| (N-T);\draw[future] (N-T) -| (N-T-);
\draw[-latex,future,nproc] ($(split-N)!.5!(mha-N-self)$) |- ($(mha-N-self.south)+(0,-.1)$) -| ($(mha-N-self)!.45!(mha-N-cross)$);
\draw[-latex,future,nproc] ($(mha-N-self)!.55!(mha-N-cross)$) |- ($(mha-N-cross.south)+(0,-.1)$) -| ($(mha-N-cross)!.45!(merge-N)$);
\draw[-latex,future] (N-T-) -- (split-T);\draw[-latex,future,tproc] (split-T) -- (mha-T);\draw[-latex,future,tproc] (mha-T) -- (merge-T);\draw[-latex,future] (merge-T) -- (out);
\draw[-latex,future,tproc] ($(split-T)!.5!(mha-T)$) |- ($(mha-T.south)+(0,-.1)$) -| ($(mha-T)!.25!(merge-T)$);
\end{scope}
\draw[defbox] ($(bbox.north west)+(.1,.3)$) rectangle ($(bbox.south east)+(-.1,-.3)$);
\node[cartouche] at (bbox.north) {\sc dec-layer$_\ell$};

\end{tikzpicture}
\end{subfigure}
\caption{\label{fig:model}The overall architecture of our ADN model (top) is a traditional Encoder-Decoder network. Each encoder and decoder layer (bottom) consists of multihead attention blocks {\sc mha} (the masked version is indicated by a masked corner), with one input (self-attention) or two inputs (cross-attention, indicated by a thicker outline). The model alternates attention in the temporal dimension ({\sc mha}$^\textrm{(T)}$), and in the spatial dimension ({\sc mha}$^\textrm{(N)}$). This is achieved by simple manipulation of indices, alternatively blending into the batch dimension $B$ the spatial dimension $N$ ({\sc split}$^\textrm{(N)}$) and the temporal dimension $T$ ({\sc split}$^\textrm{(T)}$), then restoring it ({\sc merge}$^\textrm{(N)}$ and {\sc merge}$^\textrm{(T)}$, respectively). $D$ is the embedding dimension. Skip connections are shown, but normalisation, dropout and feed-forward blocks have been omitted for clarity. All the other blocks in the figure are detailed in the text.}
\end{center}
\end{figure*}
\section{Introduction}
We are interested in the problem of prediction in multi-variate time series where nothing is known about the process generating the data. We consider time series which are structured, i.e. their measurements consist of homogeneous vectors taken at different instants in time and different points of an abstract space, called {\em locations}. Instant-location pairs are called {\em events}. This configuration is quite common in applications, such as the prediction of service demand~\cite{salinas_high-dimensional_2019}, of stock prices~\cite{elliot_time_2017}, of geophysical phenomena like rainfall~\cite{shi_convolutional_2015}, or of road traffic~\cite{li_diffusion_2017}.

In its most general formulation, the problem we address is that of predicting the measurements at a given set of (target) events, given measurements at a related set of (source) events. How source and target are related is often left implicit, but in any case, it is useful to represent each problem instance as an event grid, the Cartesian product of a set of instants and a set of locations, where the source events in the grid have known measurements while those of the target events are to be estimated. In other words, we view prediction as a form of matrix completion.

However, classical matrix completion methods work on one large incomplete matrix which they propose to complete by exploiting patterns discovered on its known parts. Instead, in this work, we focus on so-called ``short term'' prediction, where a large training set of small complete instances, from which patterns can be learnt, is used in the completion of a set of test instances, assumed sampled from the same distribution. That problem setting could be characterised as a form of ``localised'' matrix completion, where the large global matrix is only accessible through selected small local patches (the instances).

In this setting, we address the question of how to deal with prior knowledge. Prior information attached to individual events is easily dealt with, assuming it is homogeneous: it can be appended to the measurement at each event, and used as context. But relational information involving event pairs (or tuples), which we call structural prior, poses difficulties. It is tempting to explicitly incorporate it in the information diffusion mechanism of the model, but that should be done in such a way that the effect of the prior wanes when more data becomes available for training, so that eventually, the model is entirely data-driven (as in Bayesian learning). Unfortunately, this is not the case of many published models, in which the information diffusion mechanism of the model is organically impacted by the structural prior, for example through a GCN, whatever the amount of data it sees. Furthermore, it is often the case that the type or intensity of the relations captured by the structural prior differ from one domain to another, making domain adaptation more difficult if the model depends on it.

For these reasons, we propose the simplest possible model, called ADN for Attention Diffusion Network, which does not rely on any structural prior whatsoever. We choose an attention based encoder-decoder architecture, where attention is adapted to the bi-dimensionality of events as location-instant pairs. It allows information diffusion in both the temporal and spatial dimension (but not simultaneously), controlled by attention parameters entirely learnt from the data. We experimentally measure the impact of this design choice on the performance, and show that, at least in our use case, very little data is sufficient to be on a par with state of the art models which do rely on a structural prior. We then proceed to explore the advantages of our light-weight approach when dealing with missing data or domain adaptation. Missing data is simulated by randomly removing locations in the training instances, and the question is how that impacts the performance. For domain adaptation, the question is how to transfer a model learnt on one domain to a new domain with completely new locations (but the same patterns of instants), using only a small amount of data from the new domain.

\paragraph{Use case}
To fix ideas, we illustrate our proposal with the problem of data driven short term road traffic prediction, for which numerous datasets are publicly available, as well as an abundant literature using them. The event locations in that context are a set of sensor stations spread across a road network, equipped with any variety of sensors fused to produce, at regular intervals, a fixed set of features (such as vehicle speed or density) describing the traffic.
\section{Related work}
One of the first models explicitly dealing with short term prediction as spatio-temporal diffusion is DCRNN~\cite{li_diffusion_2017}, developed in the context of road traffic prediction. It is based on a RNN architecture for temporal diffusion, coupled with a GCN~\cite{defferrard_convolutional_2016,kipf_semi-supervised_2016} for spatial diffusion. They use the GRU variant of RNN, but a similar approach based on LSTM was proposed in~\cite{shi_convolutional_2015}. Some approaches~\cite{song_spatial-temporal_2020,li_spatial-temporal_2021} do not rely on RNN for temporal diffusion and instead model all diffusion as a Graph Neural Network. Several papers, while sticking to the RNN framework underlying DCRNN, expand the capacity of its GCN in various ways~\cite{lee_ddp-gcn_2020,yu_forecasting_2020,wu_graph_2019,he_stcnn:_2019}. Some papers introduce forms of attention within the RNN framework~\cite{he_stann:_2019,liu_traffic_2019}. More recent papers entirely replace the RNN framework by Transformer attention~\cite{vaswani_attention_2017}. In particular, \cite{zheng_gman_2019,xu_spatial-temporal_2020,park_st-grat_2020,cai_traffic_2020} embed temporal and spatial attention mechanisms into ad-hoc processing pipelines.

Unfortunately, all these proposals assume the existence of a graph which captures prior knowledge on the diffusion capacity between locations, or try to infer one as in the ``self-adaptive adjacency matrices'' of~\cite{wu_graph_2019,kong_stgat_2020}. On the contrary, our model is free from such assumption, and still achieves performance on a par with most single-graph based models. In effect, our model relies on the many latent graphs created by the attention heads, which are specialised for each instance and adapted to each phase of the prediction. This greatly simplifies the design of the model and increases its versatility, facilitating in particular domain adaptation (e.g. in the traffic prediction task, transferring a model learnt on the road network of one city to another).
\section{The ADN Model}
\subsection{Data model}
\label{sec:data-model}
A problem instance involves a number $N$ of locations and $T^o$ of instants. Measurement vectors, given at some of these location-instant pairs (events), are homogeneous of fixed dimension $P$. While $P$ is constant over all the instances, $N$ and $T^o$ can be instance dependent. We assume that instances are assembled in such a way that the locations in an instance are all somehow related, and for each location, measurement vectors are given at all the instants in a time interval containing a distinguished reference instant (the same for all the locations in the instance). This reflects the task, which is to predict, at each location, the measurement vectors at instants after the reference instant, given those at the reference instant and before in the interval.

Using padding and masking of instants on the right and on the left of the intervals where measurements are available (one per location), they can be made identical for all the locations in the instance, of common length $T^o{=}T{+}T'$ with $T$ instants up to (and including) the reference one, and $T'$ instants after it. Thus, a problem instance can be represented as a (masked) tensor of shape $\tuple{N,T{+}T',P}$. Using more padding and masking, $B$ such instances can be batched together in such a way they all share the same $N,T,T'$. The reference instants of the instances in the batch are aligned at position $T$ but do not need to be the same. As for locations, they need neither be aligned nor shared at all. Thus, a batch is represented as a (masked) tensor of size $\tuple{B,N,T{+}T',P}$ where each index $b$ is associated with an instance, each index pair $b,n$ with a location, each index pair $b,t$ with an instant (the reference instant of instance $b$ being at $t{=}T$) and each index $p$ with a feature of the measurement vectors.

The data contains two auxiliary pieces of information in addition to the batches: descriptors for locations and instants. They capture the prior information, assumed homogeneous in each case, available on locations and instants, and even possibly events (ignored here for simplicity). The auxiliary information about locations contains at least their identity, which carries strictly no implicit relational information on the locations themselves but allows to relate occurrences of the same location across instances. As for instants, their identity alone is a poor signal, since test instances typically involve instants never seen in training instances (usually, the train-test data partition is based on chronology, where training instances are chosen before a given date and test ones afterwards). Instead, we add information capturing the inherent periodicity of the data, assumed known or discovered by Fourier analysis. For example, in the case of road traffic prediction, the 7-day and 24-hr periodicity of traffic is captured by attaching to each instant the identity of its day of the week among the 7 days of a week and of the 5 min slot containing it among the 288 such slots in a day. Note that this does not carry any implicit relational information on instants: the system treats Monday or Tuesday as meaningless identifiers, and does not know a priori that one is just before the other, nor, similarly, that the 9:05am-9:10am slot is just before the 9:10am-9:15am slot.
\subsection{Prediction model}
Figure~\ref{fig:model} gives an overview of our prediction model, based on a traditional Encoder-Decoder architecture. The encoder and decoder pipelines each consists of an initialisation layer followed by a sequence of transformation layers. A final layer produces the prediction for each measurement feature of each target event ({\sc pred} block). In general, it is a distribution from which the actual value is sampled ({\sc sample} block), but this includes the limit case of a distribution concentrated on a single value, where sampling is trivial. During training, no sampling is done and the output of {\sc pred} is directly compared to the original input batch through some {\sc loss} function block, typically Cross-Entropy, which encompasses, in the limit case, the Mean Squared Error (MSE) and Mean Absolute Error (MAE)\footnote{MSE (resp. MAE) is the limit of a cross-entropy loss with a Normal (resp. Laplace) distribution whose variance tends to $0$.}.

In the initialisation layer {\sc enc-ini}, for each of the $B$ instances, each of the $N$ locations and each of the $T$ source instants (up to the reference one) involved in the batch, the associated event descriptors, for the instant and for the location, are retrieved and projected into the embedding dimension $D$ then summed and augmented with the positional encoding in the (chronologically ordered) sequence of instants. The result is an embedding of all the events in the batch as a tensor of shape $\tuple{B,N,T,D}$, which is then added to a projection in the embedding space of the data batch at the same source instants (up to $T$), yielding the full batch embedding of shape $\tuple{B,N,T,D}$ (in blue in Figure~\ref{fig:model}). Similarly, block {\sc dec-ini} yields batch embeddings of shape $\tuple{B,N,T',D}$ (green in the figure), collected from the reference instant $T$ up to the penultimate one $T{+}T'{-}1$, i.e. the target instants, shifted to the right, as in the standard Transformer model of~\cite{vaswani_attention_2017}.

Now, in the standard Transformer model, the transformation layers {\sc enc-layer}$_\ell$ and {\sc dec-layer}$_\ell$ manipulate 3-way tensors of shape $\tuple{B,S,D}$ or $\tuple{B,S',D}$ , where $B$ is the batch size, $S,S'$ the source and target sequence lengths and $D$ the embedding dimension of the model. The associated attention matrix in each attention head is then of size $S^2$, $S'^2$ or $SS'$, depending on whether it is a self- or cross- attention, in the encoder or the decoder.

In our model, on the other hand, the transformation layers have to deal with 4-way tensors. For example, the encoder self-attention has to deal with tensors of shape $\tuple{B,N,T,D}$. They could of course be reshaped into 3-way tensors of shape $\tuple{B,NT,D}$, which would amount to treating events atomically, ignoring their bi-dimensionality as location-instant pairs. They would then be amenable to the standard Transformer treatment of sets of cardinality $NT$, with self-attention matrices of size $(NT)^2$, which is usually unfeasible. We therefore adopt a simple dimensionality reduction technique, known for a long time in the world of convolution networks under the name ``spatial separability''~\cite{szegedy_rethinking_2015}. Since attention is a generalised form of convolution~\cite{andreoli_convolution_2019}, separability works all the same with attention, where it is also known as axial attention~\cite{ho_axial_2019}.

Separable attention processes event-indexed objects along each dimension of the events (temporal and spatial) separately and alternately, just as spatially separable convolutions alternate processing images along their width and height dimensions. While the ``atomic event'' approach with attention matrices of size $(NT)^2$ would allow each event to directly influence all the other events in the matrices, with separable attention, the influence is indirect: an event $n,t$ can influence an event $n',t'$ first through the influence of $t$ on $t'$ at location $n$ in the temporal attention, then the influence of $n$ on $n'$ at instant $t'$ in the spatial attention. The presence of multiple transformation layers in the pipeline ensures that the two types of influences are thoroughly interleaved, and the order in which the temporal and spatial attention are performed within each layer ends up being unimportant. For the sake of design simplicity, we depart here from models which try to make that order instance dependent and learnable by adding to each layer a gating mechanism arbitrating between spatial and temporal attention~\cite{wu_graph_2019,kong_stgat_2020}.

The obvious advantage of separable attention is that the attention matrices are now only of size $N^2$ and $T^2$, respectively, assumed to be feasible, or at least better amenable to the many proposals in the literature aiming at breaking the quadratic space complexity of attention~\cite{tay_efficient_2020}. In practice, separability is straightforward to implement, as it amounts to alternately reshaping each batch of shape $\tuple{B,N,T,D}$ as shown in Figure~\ref{fig:model}, into 3-way tensors of shape $\tuple{BN,T,D}$ (dashed line) then $\tuple{BT,N,D}$ (dotted line), each processed by standard Transformer attention blocks. Of course, separability as described above for the encoder self-attention is similarly applied to the other attention blocks in the model (decoder self- and cross-attention).

We use teacher forcing~\cite{williams_learning_1989}, as in Transformer. At train time, the model is used as an auto-encoder: when presented with a batch, the model is applied to it just once, and the loss measures the difference between output and input. At test time, the model is used as an auto-regressor: when presented with a batch, the model is iteratively applied, with the output of an iteration passed as input to the next.
\section{Experiments}
Beyond reporting the performance of our model on benchmarks, we conduct a number of experiments measuring the impact of: (i) data scarcity; (ii) missing data; (iii) large instances. We further analyse the results of our experiments to try and explain some of the performances of our model. Finally, we consider the problem of domain adaptation.
\subsection{Experimental setting}
\subsubsection{Datasets}
Experiments are conducted on three public traffic datasets PEMS-BAY, METR-LA and PEMS07 released by \cite{li_diffusion_2017} and \cite{song_spatial-temporal_2020}. The first two are the most commonly used for measuring model performances, and consist of 207 and 325 locations, respectively. We also tested our model on PEMS07 (883 locations), to check its scalability to larger road traffic networks.
All three datasets provide data as speed averages or sum of traffic flow on contiguous 5 minute intervals. They also include distance matrices, holding inter-location distances measured on the road network. These are used only in the baselines, not in our model (ADN) since they constitute precisely the kind of structural prior which we seek to avoid. Instances in each of the three datasets are obtained by splitting the data into 2 hour moving windows, with one-hour overlap. Each instance thus consists of 24 instants (one every 5 minutes) and the reference instant is always taken to be the middle one (at position 12). We follow the same procedure for dispatching instances into training, validation, and test subsets as in the literature (details provided in Appendix). Although some authors do data cleaning (like excluding zeros), we do not perform any preprocessing of the data.
\subsubsection{Hyper-parameters and implementation}
The model hyperparameters are tuned on the PEMS-BAY dataset and reused on all the other datasets. The best results are achieved with a model composed of 3 layers, both in the encoder and the decoder, each built with 2-head spatial attentions and 4-head temporal attentions. Model embedding size $D$: 32; dimension of the feed-forward layer: 256; dropout value: 0.3.

All models are implemented using the PyTorch machine learning framework and trained on a machine with 4 NVIDIA V100 GPUs with 32GB memory per GPU. For all datasets, we set the batch size to 64 and limit the training to 100 epochs. Models are trained using the MAE loss (Mean Absolute Error, or L1) with the Adam optimizer with parameters $\beta_1{=}0.9$, $\beta_2{=}0.98$ and $\epsilon{=}1e^{-9}$. The initial learning rate is set to $0.002$ and halved at the 15th, 30th and 45th epoch. Training gradients are clipped at $0.1$.
\subsection{Performance}
\subsubsection{Base performance}
\begin{table*}
    \centering
    \begin{tabular}{l | r | c c | c c c c c c | c}
          & & HA & ARIMA & FNN & STGCN & SLCNN & DCRNN & GWN & STGAT & ADN$^{\textrm{Ours}}$ \\
          \hline
          \multirow{3}{*}{\rotatebox[origin=c]{90}{\parbox[c]{1cm}{\centering PEMS-BAY}}} & 15 min & 2.88 & 1.62 & 2.20 & 1.36 & 1.44 & 1.38 & \textbf{1.30} & \em{1.32} & \textbf{1.30} \\
          & 30 min & 2.88 & 2.33 & 2.30 & 1.81 & 1.72 & 1.74 & 1.63 & \textbf{1.61} & \em{1.62}\\
          & 1 hour & 2.88 & 3.38 & 2.46 & 2.49 & 2.03 & 2.07 & 1.95 & \em{1.91} & \textbf{1.90}\\
         \hline
         \hline
          \multirow{3}{*}{\rotatebox[origin=c]{90}{\parbox[c]{1cm}{\centering METR-LA}}}
          & 15 min & 4.16 & 3.99 & 2.20 & 2.88 & \textbf{2.53} & 2.77 & 2.69 & 2.66 & \em{2.61} \\
          & 30 min & 4.16 & 5.15 & 4.23 & 3.47 & \textbf{2.88} & 3.15 & 3.07 & 3.01 & \em{2.98} \\
          & 1 hour & 4.16 & 6.90 & 4.49 & 4.59 & \textbf{3.30} & 3.60 & 3.53 & 3.46 & \em{3.42} \\
         \hline
    \end{tabular}
    \caption{\label{tab:results}MAE at various horizons on PEMS-BAY and METR-LA datasets. All the baseline results are taken from the corresponding papers. In all cases, our model performs either best (bold) or second best (italics).}
\end{table*}
Here, we compare our model on the PEMS-BAY and METR-LA benchmarks with two types of baselines: traditional methods like Historical Average (HA), and Auto-Regressive Integrated  Moving Average (ARIMA) with Kalman filter; and neural methods: simple Feed-Forward Neural Network (FNN, two hidden layers and L2 regularization), Spatio-Temporal Graph Convolutional Networks (STGCN) \cite{yu_spatio-temporal_2018}, Spatio-Temporal Graph Structure Learning for Traffic Forecasting (SLCNN) \cite{zhang_spatio-temporal_2020}, Diffusion Convolutional Recurrent Neural Network (DCRNN) \cite{li_diffusion_2017}, Graph WaveNet (GWN) \cite{wu_graph_2019} and Spatial-Temporal Graph Attention Networks for Traffic Flow Forecasting (STGAT) \cite{kong_stgat_2020}. For the evaluation, we report Mean Absolute Errors (MAE), Root Mean Squared Errors (RMSE) and Mean Absolute Percentage Error (MAPE) and performances are compared on predictions at 15, 30 and 60 minute horizons. Missing values are excluded in calculating these metrics. Table~\ref{tab:results} shows only the MAE metric, while the full results are provided in Appendix. Our model is on a par or outperforms the baselines (except, surprisingly, SLCNN on just the METR-LA dataset, but recall our hyper-parameters were optimised on PEMS-BAY only).
\subsubsection{Impact of data scarcity}
\begin{figure}
\begin{center}
\includegraphics[scale=0.4]{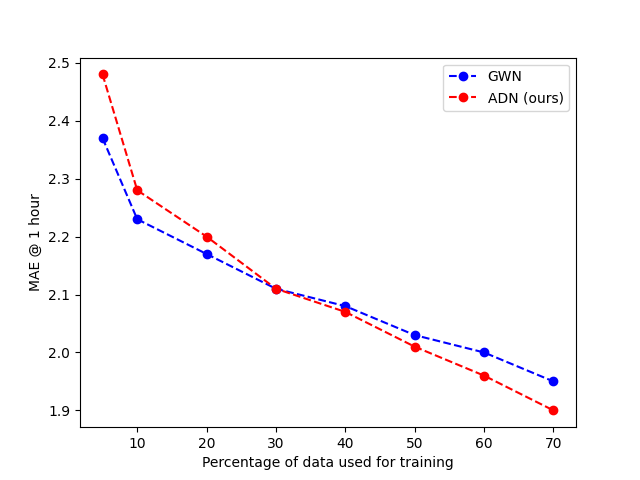}
\end{center}
\caption{\label{fig:scarce-data}PEMS-BAY with (simulated) scarce data. The structural prior gives a performance advantage to the GWN baseline at very scarce data regime, but becomes a hindrance when data is abundant.}
\end{figure}
As might be expected, the fact that our model does not rely on a structural prior may affect performance when data is scarce, since, in that case, limited training is not compensated by the prior. We put that assumption to the test, by artificially reducing the training dataset of PEMS-BAY and observing the degradation of performances, on our model and on one of the baselines which do use a structural prior, namely GWN~\cite{wu_graph_2019}. We do not reduce the number of training locations, so that the effect of the structural prior is unchanged, but we reduce the number of training instants: we remove from the training set the instances whose reference instant is outside the $x$ percent earliest, while keeping the same validation and test sets (which are themselves chosen as having the latest reference instants). The results for various values of $x$, in Figure~\ref{fig:scarce-data}, show that the performance advantage brought by the structural prior remains limited overall, even at scarce data regimes, at least above 5\%\footnote{Observe that below 1 week of data, some days of the week are not even represented in the training data, but still present in test, deeply impacting the performance.}. Furthermore, not only does this advantage tend to disappear with more data, but, beyond a certain threshold (ca. 30\%), the effect becomes counter-productive and the prior-free model (ours) starts to outperform the model with prior. That confirms our intuition that the prior does not vanish with more data (GWN is not a Bayesian model), but keeps affecting the performance even when the contribution of data alone would be superior.
\subsubsection{Impact of large instances}
\begin{table*}
    \centering
    \begin{tabular}{l || c  c  c  c  c  c | c }
         & FC-LSTM & DCRNN & STGCN & GWN & STSGCN & STFGNN & ADN$^{\textrm{Ours}}$ \\
         \hline
         PEMS07 & 29.98 $\pm$ 0.42 & 25.30 $\pm$ 0.52 & 25.38 $\pm$ 0.49 & 26.85 $\pm$ 0.05 & 24.26 $\pm$ 0.14 & 22.07 $\pm$ 0.11 & \textbf{21.62 $\pm$ 0.10} \\
        \hline
    \end{tabular}
    \caption{\label{fig:pems07-results}MAE at 1 hour on PEMS07. All the baseline results are taken from \cite{li_spatial-temporal_2021}.}
\end{table*}
One limitation of our model, and more generally of all attention-based models, is the quadratic space requirement of attention matrices. While temporal attention poses no problem since the temporal dimensions of instances remain very small ($T{=}T'{=}12$), spatial attention may become a problem as the possible number $N$ of locations in an instance increases, for example when tackling the PEMS07 dataset, which has the largest number of locations (more than double that of the other datasets).

As expected, with PEMS07, spatial attention matrices do not fit into memory on our standard hardware. But, taking advantage of the flexibility of our model, it is possible to work around this problem and even outperform other models. Indeed, while it may seem desirable to deal with instances involving all the locations in the network, this is in no way compulsory, and we can slice the instances along the spatial dimension as much as needed to reduce the size of the slices. We don't have to worry about slicing the corresponding structural prior since we don't make use of one.

If for example we partition the location set into $k$ subsets, one full instance of spatial dimension $N$ becomes $k$ partial instances of dimension approximately $\frac{N}{k}$ (the dimension of the slices need not be strictly equal, nor even close). Again, as we preclude the use of a structural prior, we do not seek to optimise the way partitioning is done, e.g. along geographical lines as in~\cite{mallick_graph-partitioning-based_2020}, and instead draw fully random partitions. Actually, partitions are redrawn at each training epoch. Our experiments show that the best results are reached by partitioning locations into $k{=}16$ subsets during the training, which allows the attention matrices to all fit in memory.

Performance of this approach is compared to Spatial-Temporal Fusion Graph Neural Networks (STFGNN)~\cite{li_spatial-temporal_2021}, for which results on the PEMS07 dataset are available, as well as all the baselines presented in that paper: Long Short-Term Memory - Fully Connected Neural Network (FC-LSTM), DCRNN, STGCN, GWN, Spatial-Temporal Synchronous Graph Convolutional Networks (STSGCN) \cite{song_spatial-temporal_2020}. Following \cite{li_spatial-temporal_2021}, we run experiments 10 times and report the mean and standard deviation in Table~\ref{fig:pems07-results}.
\subsubsection{Impact of missing data}
\begin{figure}
\begin{center}
\includegraphics[scale=0.4]{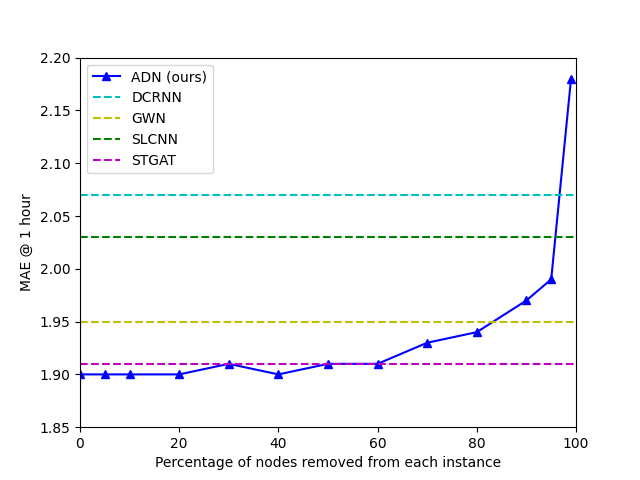}
\end{center}
\caption{\label{fig:missing-data}PEMS-BAY with (simulated) missing data. The dashed lines correspond to baseline models learnt with the full dataset, shown here for reference.}
\end{figure}
Often, datasets are not of uniform quality, and some instances may contain ``holes'', typically if the sensors at some locations fail and report no measurement during the time span of the instance. In our model, such locations are simply removed from the instance, and again, we don't have to worry about applying the same removal in the structural prior.

We conducted an experiment on the PEMS-BAY dataset, simulating missing data by removing at random a proportion $x$ of locations in each instance. The results for various values of $x$, in Figure~\ref{fig:missing-data}, show the remarkable robustness of our model to missing locations in the training set. The impact on performance is negligible at levels up to 60\% missing locations, and remains competitive at up to 95\% missing locations. Of course, it is important to observe that random location removal is performed independently in each instance: the same results could not be achieved if removals were inter-dependent (e.g., obviously, in the extreme case where the random subset of removed locations were constrained to be the same in all instances).
\subsection{Analysis of the results}
\subsubsection{Spatial analysis}
\begin{figure}
\begin{center}
\includegraphics[scale=0.4]{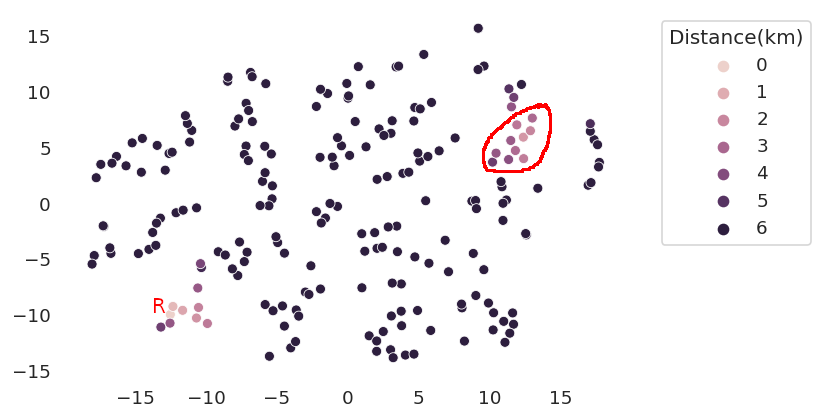}
\end{center}
\caption{\label{fig:location-embeddings}Visualisation of the PEMS-BAY location embeddings. The colours indicate the distance, according to the prior, to the distinguished location labelled {\tt R}.}
\end{figure}
To further investigate the effect of using a structural prior (in the baselines) vs not using it (in our model), we sought to compare the latent relationship between locations, as built by our model, and the explicit relationship provided by the prior, based on the inter-location distances on the road network.

In fact, our model builds many relationships between locations, essentially one for each instance and in each head of each attention block (examples are shown in Appendix). But they all indirectly derive from the embeddings of the location descriptors learnt in the initial layer of the encoder ({\sc enc-ini} in Figure~\ref{fig:model}). Figure~\ref{fig:location-embeddings} gives a 2D representation of these embeddings (of dimension $D$) obtained by a standard statistical dimensionality reduction method, t-SNE~\cite{maaten_visualizing_2008}. The figure shows distinct clusters which, after verification, correspond roughly to distinct segments of the road network. In other words, the relational information constructed by our model is close to that provided by the structural prior.

However, we noticed some discrepancies. Take location labelled {\tt R} (in red) on the figure. We coloured all the other locations by their distance to {\tt R} according to the structural prior (the darker the colour, the longer the distance). We would expect that locations which are close in the prior are also close in the model, but we observe that a cluster of locations close to {\tt R} in the prior (i.e. with light colour), encircled in red in the figure, appear far from {\tt R} in the model. Taking a closer look at these exceptions, we realised they correspond to locations which are indeed close to {\tt R} on the road network, but only through a path which is very unlikely, e.g. involving a U-turn between the lanes in opposite directions of a motorway, through a bridge over it.

In all the datasets we use, the prior is obtained as the inter-location network distances as computed by some mapping tool such as OpenStreetMap, i.e. the length of the shortest path between locations on the road network, but that does not take into account the probability of vehicles following that path. This is the kind of defects of the prior which can have a lasting detrimental effect on models which rely on a prior but do not ensure that it gives way to data when data becomes abundant. We suspect that is what happens in most baselines, which use a prior but not in a Bayesian way. Our model, which does not involve a prior at all, is safe in that respect.
\subsubsection{Temporal analysis}
\begin{figure}
\begin{center}
\includegraphics[scale=0.35]{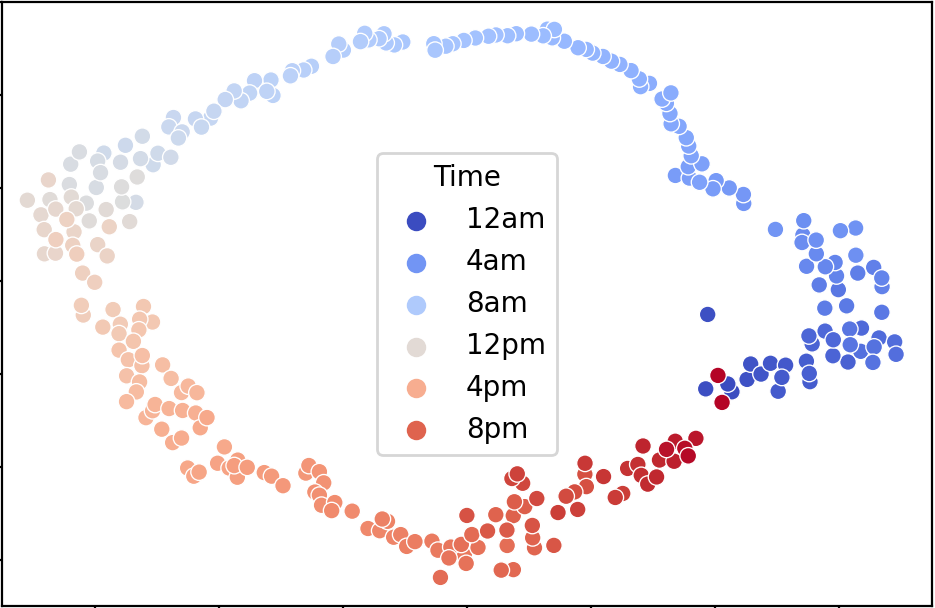}
\includegraphics[scale=0.35]{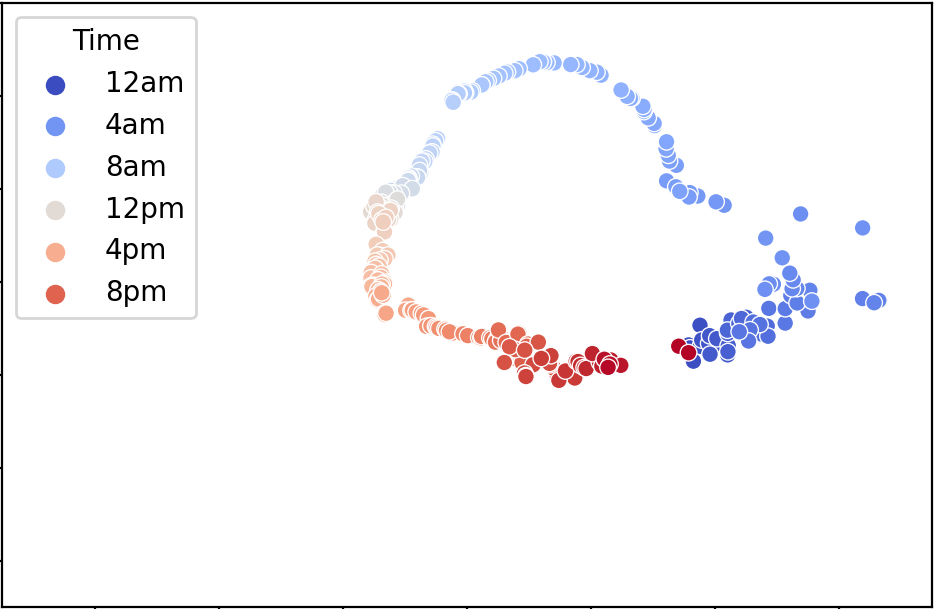}
\end{center}
\caption{\label{fig:instant-embeddings}Visualisation of the embeddings of the 288 time slots of each day, on Friday (left) and Saturday (right), in the same referential given by the bounding box.}
\end{figure}
We conducted a similar visual analysis of the embeddings on instants, rather than locations. A t-SNE representation of the different time slots in the different days of the week is given in Figure~\ref{fig:instant-embeddings}, showing only Friday (Monday to Thursday are similar) and Saturday (Sunday is similar). It shows that our model does learn quite precisely the ordering of time slots in a day, as well as their cyclic nature, although this information is not contained in the instant descriptors given to the system, limited to undistinguishability, not ordering information. While it might be argued that positional encoding, applied in the {\sc enc-ini} and {\sc dec-ini} blocks of the model, may carry implicit ordering, this is not what underlies the results of Figure~\ref{fig:instant-embeddings}, since all the experiments reported in this paper have been conducted with no positional encoding! In fact, the only place in the model where explicit temporal ordering information is indeed exploited is in the mask of the decoder self-attention, which prevents instant embeddings in one layer from influencing the embeddings of the explicitly designated {\em previous} instants in the next layer.
\subsection{Domain adaptation}
\begin{figure}
\begin{center}
\includegraphics[scale=0.3]{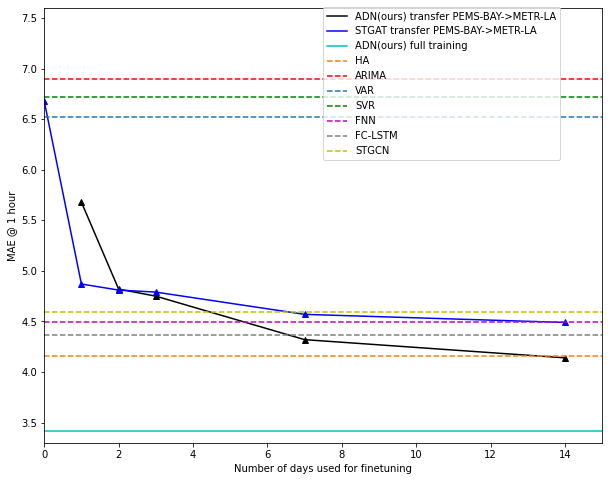}
\end{center}
\caption{\label{fig:domain-adaptation}Domain adaptation from PEMS-BAY to METR-LA. The dashed lines correspond to baseline models learnt on the full target domain, shown here for reference. Our model learnt on the full target domain is also reported.}
\end{figure}
Finally, we consider the problem of domain adaptation, where a model learnt with data from a source domain is applied to a target domain. To simplify, we focus on the case where the source and target domains differ only in their locations, and not in their periodicity patterns (which fully characterize the instants), although these could also be in the scope of the adaptation. More precisely, we assume that source and target domains have no location in common, as in the case of PEMS-BAY and METR-LA already considered in~\cite{kong_stgat_2020}. Using different datasets,~\cite{mallick_transfer_2020} consider a similar problem under the name ``transfer learning'' (it is unclear how that differs from what is discussed here).

Models which rely on a structural prior seem to have an advantage when it comes to domain adaptation: they can leverage the prior to do zero-shot adaptation, i.e. without seeing any data from the new domain, just the prior. However, it can be expected that the quality of the prediction in that case will be rather poor and probably of no practical use. Furthermore, while in the traffic application the structural prior is cheap compared to data acquisition, it may not be the case in other contexts. Instead, we propose a ``few shot'' adaptation approach, where data over a very short period (a few days vs several months for the full dataset) is accessible from the target domain for fine-tuning the model learnt on the source domain.

In fact, only a small part of the model needs fine-tuning. Indeed, the parameters of the ADN model consist of two sets:
\begin{itemize}
\item the parameters of the attention blocks in the encoder and decoder: we may assume these are generic, and characterize the road traffic prediction task per se, with no reference to a particular domain, so they do not need fine-tuning;
\item the initial embeddings of the events (locations and instants): we assumed similar periodicity patterns between source and target, so the instant embeddings can be reused as such, while nothing can be reused from the source location embeddings since each location is entirely characterised by its identity, and no identity is shared between source and target domains.
\end{itemize}
Hence, in the fine-tuning step, we only need to learn, from scratch, the embeddings of the target domain locations, while keeping all the other parameters fixed.

Results for different amounts of fine-tuning data, expressed as the number of days its instances span, are presented in Figure \ref{fig:domain-adaptation}. We compare performance with the STGAT model~\cite{kong_stgat_2020}. To be fair, instead of comparing with just the number reported in that paper, which corresponds to zero-shot learning by substitution of the prior (no fine-tuning), we do both a substitution of the prior and a fine-tuning of (all) the parameters of the STGAT model on the same data. In both models, we run 100 epochs of fine-tuning using the hyper-parameters learnt on the source domain PEMS-BAY.

The experimental results show that our model needs only a small amount of fine-tuning data to reach good performance: at 1hr horizon, two days are enough to cancel the performance advantage the structural prior gives to STGAT. Beyond that amount, our model keeps improving, at a faster rate than STGAT. More details are available in Appendix.
\section{Conclusion}
In this paper, we have considered the problem of structured time-series prediction, and proposed a light-weight model, not relying on any prior knowledge on the relations between events in the time series. We show that such structural prior, if used in a non-Bayesian way as in most models from the literature, can have a detrimental effect on performance. We also show that our model, being more versatile, can better deal with the problems of missing data and domain transfer.

\clearpage

\printbibliography

\clearpage

\appendix
\section{Additional experimental results}

\begin{table}[p]
    \centering
    \begin{tabular}{l |c | c | c | c | c | c }
    Dataset & \# of nodes & Data range & Data type & Training & Validation & Testing \\
    \hline
    METR-LA & 207 & 3/1/2012 - 1/30/2012 & speed & 70\% & 10\% & 20\% \\
    \hline
    PEMS-BAY & 325 & 1/1/2017 - 5/3/2017 & speed & 70\% & 10\% & 20\%  \\
    \hline
    PEMS07 & 883 & 5/1/2017 - 8/31/2017 & flow & 60\% & 20\% & 20\% \\
    \hline
    \end{tabular}
    \caption{\label{tab:datasets}Details of the size and nature of the 3 datasets used in the experimental part of the paper.}
\end{table}

\begin{table}
    \centering
    \begin{tabular}{c || c | c | c || c | c | c }
         & \multicolumn{3}{c||}{PEMS-BAY (MAE, RMSE, MAPE\%)} & \multicolumn{3}{|c}{METR-LA (MAE, RMSE, MAPE\%)}\\
         \hline
         \hline
         Model & 15 min & 30 min & 60 min & 15 min & 30 min & 60 min \\
         \hline
         HA & 2.88  5.59  6.8 & 2.88  5.59  6.8 & 2.88  5.59  6.8 & 4.16  7.80  13.0 & 4.16  \: 7.80  13.0 & 4.16  \: 7.80  13.0 \\
         ARIMA & 1.62  3.30  3.5 & 2.33  4.76  5.4 & 3.38  6.50  8.3 & 3.99  8.21  \: 9.6 & 5.15  10.45  12.7 & 6.90  13.23  17.4 \\
         \hline
         FNN & 2.20  4.42  5.2 & 2.30  4.63  5.4 & 2.46  4.98  5.9 & 3.99  7.94  \: 9.9 & 4.23  \: 8.17  12.9 & 4.49  \: 8.69  14.0 \\
         STGCN & 1.36  2.96  2.9 & 1.81  4.27  4.2 & 2.49  5.69  5.8 & 2.88  5.74  \: 7.6 & 3.47  \: 7.24  \: 9.6 & 4.59  \: 9.40  12.7 \\
         SLCNN & 1.44  2.90  3.0 & 1.72  3.81  3.9 & 2.03  4.53  4.8 & \textbf{2.53}  5.18  \: \textbf{6.7} & \textbf{2.88}  \: \textit{6.15}  \: \textbf{8.0} & \textbf{3.30}  \: \textit{7.20}  \: \textbf{9.7}\\
         DCRNN & 1.38  2.95  2.9 & 1.74  3.97  3.9 & 2.07  4.74  4.9 & 2.77  5.38  \: 7.3 & 3.15  \: 6.45  \: 8.8 & 3.60  \: 7.59  10.5 \\
         GWN & \textbf{1.30}  \textbf{2.74}  \textbf{2.7} & 1.63  \textit{3.70} \textbf{3.7} & 1.95  \textit{4.52}  \textit{4.6} & 2.69  5.15  \: \textit{6.9} & 3.07  \: 6.22  \: 8.4 & 3.53  \: 7.37  10.0 \\
         STGAT & 1.32  \textit{2.76}  2.8 & \textbf{1.61}  \textbf{3.68}  \textbf{3.7} & \textit{1.91} \textbf{4.43}  \textit{4.6} & 2.66  \textbf{5.12}  \: \textit{6.9} & 3.01  \: \textbf{6.12}  \: \textit{8.1} & 3.46  \: \textbf{7.19}  \: \textit{9.8}\\
         \hline
         ADN$^{\textrm{Ours}}$ & \textbf{1.30}  2.86  \textbf{2.7} & \textit{1.62}  3.81  \textit{3.8} & \textbf{1.90}  4.56  \textbf{4.5} & \textit{2.61}  \textit{5.14}  \: \textbf{6.7} & \textit{2.98}  \: 6.25  \: 8.2 & \textit{3.42}  \: 7.44  10.0
	     \\
    \end{tabular}
    \caption{\label{tab:results2}Complete performance results of our model on the two moderate-sized benchmark datasets, PEMS-BAY and METR-LA. Baselines are the same as in the main text, described in the section of the paper on ``Base Performance''. Results for the baseline models are taken from the corresponding papers.}
\end{table}
\begin{table}
    \centering
    \begin{tabular}{l || c | c | c | c | c | c | c}
         & FC-LSTM & DCRNN & STGCN & GWN & STSGCN & STFGNN & ADN$^{\textrm{Ours}}$ \\
         \hline
         MAE & 29.98 $\pm$ 0.42 & 25.30 $\pm$ 0.52 & 25.38 $\pm$ 0.49 & 26.85 $\pm$ 0.05 & 24.26 $\pm$ 0.14 & 22.07 $\pm$ 0.11 & \textbf{21.62 $\pm$ 0.10} \\
         MRSE & 45.94 $\pm$ 0.57 & 38.58 $\pm$ 0.70 & 38.78 $\pm$ 0.58 & 42.78 $\pm$ 0.07 & 39.03 $\pm$ 0.27 & \textbf{35.80 $\pm$ 0.18} & 37.10 $\pm$ 0.04 \\
         MAPE\% & 13.20 $\pm$ 0.53 & 11.66 $\pm$ 0.33 & 11.08 $\pm$ 0.18 & 12.12 $\pm$ 0.41 & 10.21 $\pm$ 1.65 & \: 9.21 $\pm$ 0.07 & \:  \textbf{8.93 $\pm$ 0.06} \\
         \hline
    \end{tabular}
    \caption{Complete performance results concerning the larger benchmark dataset PEMS07. Baselines are the same as in the main text, described in the section of the paper on ``Impact of large instances''. Results of the baseline models are taken from the corresponding paper.}
    \label{fig:pems07-results2}
\end{table}
\begin{table}
    \centering
    \begin{tabular}{l || c | c | c | c |}
         &  & \multicolumn{3}{c|}{MAE RMSE MAPE\%} \\
         Model & $k$-shots & 15 mins & 30 mins & 1 hour \\
         \hline
         \multirow{5}{*}{STGAT} & - & \textbf{3.96} \textbf{8.32} \textbf{9.98} & \textbf{5.14} \textbf{10.40} \textbf{12.51} & \textbf{6.68} \textbf{12.39} \textbf{18.71} \\
         & 1 day & \textbf{3.35} \textbf{6.30} \textbf{9.56} & \textbf{4.27} \: \textbf{8.20} \textbf{12.34} & \textbf{4.89} \: \textbf{9.72} \textbf{15.23}\\
         & 2 days & \textbf{3.32} \textbf{6.26} \textbf{9.12} & \textbf{4.21} \: \textbf{7.80} \textbf{12.05} & 4.82 \: 9.96 14.74\\
         & 3 days & \textbf{3.31} \textbf{6.23} \textbf{9.00} & 4.15 \: 8.68 11.55 & 4.79 \: 9.62 14.13 \\
         & 7 days & 3.23 6.18 9.13 & 3.82 \: 7.54 11.49 & 4.57 \: 8.99 14.48\\
         & 14 days & 3.17 5.90 8.93 & 3.71 \: 7.19 10.84 & 4.49 \: 8.80 13.58\\
         \hline
         \multirow{5}{*}{ADN$^{\textrm{Ours}}$} & - & - & - & - \\
         & 1 day & 3.86 8.28 10.84 & 4.72 10.13 13.61 & 5.68 11.92 16.54\\
         & 2 days & 3.59 7.57 9.91 & 4.23 \: 8.98 12.09 & \textbf{4.81} \: \textbf{9.45} \textbf{13.74}\\
         & 3 days & 3.41 7.37 9.45 & \textbf{3.96} \: \textbf{7.62} \textbf{11.38} & \textbf{4.75} \: \textbf{9.37} \textbf{13.66}\\
         & 7 days & \textbf{3.18} \textbf{6.11} \textbf{8.73} & \textbf{3.70} \: \textbf{7.36} \textbf{10.84} & \textbf{4.32} \: \textbf{8.75} \textbf{13.26}\\
         & 14 days & \textbf{3.04} \textbf{5.83} \textbf{8.13} & \textbf{3.53} \: \textbf{7.04} \textbf{10.07} & \textbf{4.14} \: \textbf{8.42} \textbf{12.38}\\

         \hline

    \end{tabular}
    \caption{\label{tab:domain-adaptation}The full results of our experiments of domain adaptation from PEMS-BAY to METR-LA.}
\end{table}

\begin{figure}[h]
    \begin{subfigure}{.5\textwidth}
    \centering
    \includegraphics[scale=0.7]{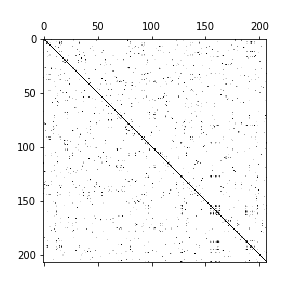}
    \end{subfigure}
    \begin{subfigure}{.5\textwidth}
    \centering
    \includegraphics[scale=0.7]{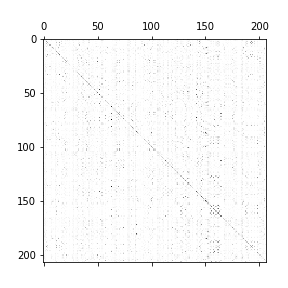}
    \end{subfigure}
  \caption{\label{fig:adj-matrix} Left: adjacency matrix for METR-LA, used as structural prior in DCRNN model; right: decoder spatial attention matrix for a randomly chosen instance in our model. Although we display just one (out of six) attention matrix, some similarities can be detected.}
\end{figure}

\begin{figure}[h]
    \begin{subfigure}{.5\textwidth}
    \centering
    \includegraphics[height=0.7\textwidth]{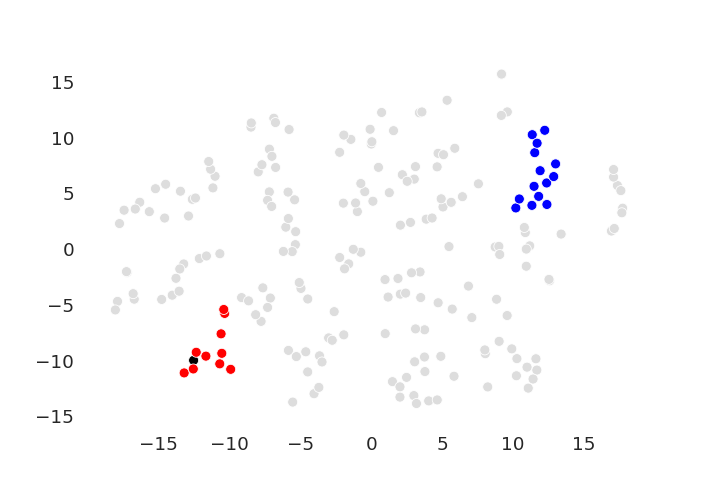}
    \end{subfigure}
    \begin{subfigure}{.5\textwidth}
    \centering
    \includegraphics[width=0.8\textwidth]{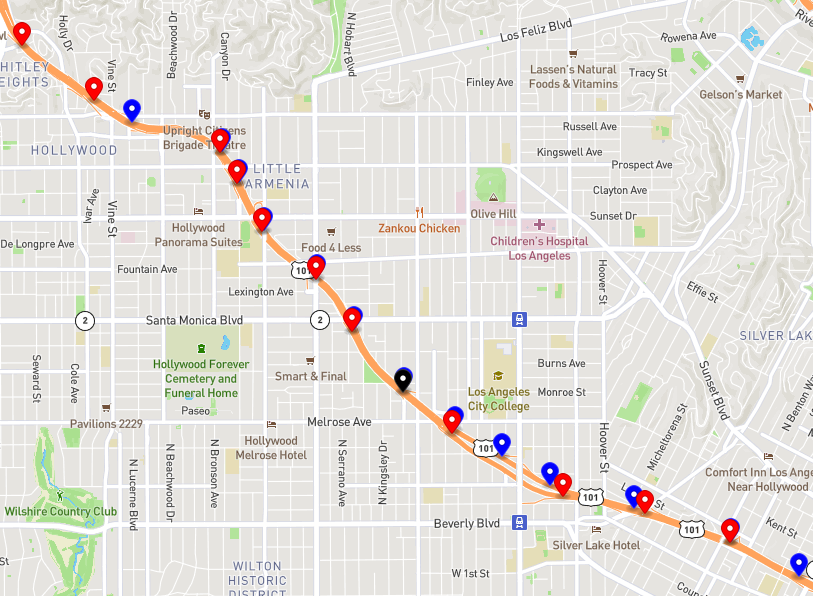}
    \end{subfigure}
    \caption{\label{fig:nodes-on-map}Correspondence between the actual position on the road network of a set of locations (right figure) and the t-SNE representation of their embeddings in the model (left). The East-West and West-East lanes of the motorway, with blue and red markers, respectively, clearly correspond to two distinct clusters in the embedding space. In the structural prior based on network distances, these two sets of locations are much closer together, since there always exist short paths allowing U-turn from one lane to the other of the motorway. Such paths are very seldom used and should have no impact on the diffusion of traffic information. Models using the structural prior will diffuse information through these unlikely paths, while our model is clearly not affected with this.}
\end{figure}

\end{document}